\documentclass{IEEE_lsens}

\usepackage{textcomp}
\usepackage{graphicx}
\usepackage[noadjust]{cite}
\usepackage[T1]{fontenc}
\usepackage{amsmath}
\usepackage{amssymb}
\usepackage[cmintegrals]{newtxmath}
\usepackage{bm}
\usepackage{array}
\usepackage{url}

\usepackage[pdftex,
            pdfusetitle,
            bookmarks=true,
            bookmarksnumbered=true,
            pdfpagemode=UseOutlines,
            colorlinks=true,
            linkcolor=black,
            citecolor=black,
            urlcolor=black]{hyperref}

\usepackage{etoolbox}
\makeatletter
\patchcmd{\section}{\vspace*{0.5\baselineskip}}{\vspace*{0.4\baselineskip}}{}{}
\patchcmd{\subsection}{\vspace*{0.5\baselineskip}}{\vspace*{0.4\baselineskip}}{}{}
\makeatother

\usepackage{stfloats}

\usepackage{tikz}
\usetikzlibrary{shapes.geometric, arrows.meta, positioning}

\tikzset{
  block/.style = {
    rectangle,
    draw=black,
    fill=blue!10,
    rounded corners,
    align=center,
    text width=3.5cm,
    minimum height=1cm
  },
  arrow/.style = {
    -{Latex[length=3mm]},
    thick
  }
}

\hypersetup{
  pdftitle={Deep Neural Networks Enable High-Accuracy Detection of Odor Presence},
  pdfsubject={Neural Networks, Odor Detection},
  pdfauthor={Author information},
  pdfkeywords={Olfactory neural signals, deep neural networks, odor detection, extracellular recordings, time-frequency analysis}
}

\begin{document}

\markboth{Vol.~1, No.~3, May~2023}{0000000}

\IEEELSENSarticlesubject{Sensor Applications}

\title{Detection of Odor Presence via Deep Neural Networks}

\author{\IEEEauthorblockN{Matin Hassanloo\IEEEauthorrefmark{1}\IEEEauthorieeemembermark{1}, Ali Zareh\IEEEauthorrefmark{1},
and~Mehmet Kemal Özdemir\IEEEauthorrefmark{2}  
}
\IEEEauthorblockA{\IEEEauthorrefmark{1}Department of Computer Engineering, Istanbul Medipol University, Kavacık Campus, Istanbul, 34810, Turkey\\
\IEEEauthorrefmark{2}Department of Electrical and Electronics Engineering, Istanbul Medipol University, Kavacık Campus, Istanbul, 34810, Turkey\\
\IEEEauthorieeemembermark{1}Member, IEEE\\
 }
\thanks{Corresponding author: Matin Hassanloo (matin.hassanloo@std.medipol.edu.tr).}
}

\IEEELSENSmanuscriptreceived{ }

\IEEEtitleabstractindextext{%
\begin{abstract}
Odor detection underpins food safety, environmental monitoring, medical diagnostics, and many more fields. The current artificial sensors developed for odor detection struggle with complex mixtures while non-invasive recordings lack reliable single-trial fidelity. To develop a general system for odor detection, in this study we present a preliminary work where we aim to test two hypotheses: (i) that spectral features of local field potentials (LFPs) are sufficient for robust single-trial odor detection and (ii) that signals from the olfactory bulb alone are adequate. To test two hypotheses, we propose an ensemble of complementary one-dimensional convolutional networks (ResCNN and AttentionCNN) that decodes the presence of odor from multichannel olfactory bulb LFPs. Tested on 2,349 trials from seven awake mice, our final ensemble model supports both hypotheses, achieving a mean accuracy of 86.6\%, an F1-score of 81.0\%, and an AUC of 0.9247, substantially outperforming previous benchmarks. In addition, the t-SNE visualization confirms that our framework captures biologically significant signatures. These findings establish the feasibility of robust single-trial detection of the presence of odor from extracellular LFPs, as well as demonstrate the potential of deep learning models to provide a deeper understanding of olfactory representations.

\end{abstract}

}

\maketitle

\section{Introduction}
The development of advanced sensor technologies for the reliable detection of odorants is a critical challenge in different fields ranging from environmental safety to medical diagnostics \cite{ref1, ref2, ref3}. Moving beyond the limitations of conventional electronic noses (e-noses), the field of neural sensing aims to emulate the olfactory system, using its rapid and complex processing to create high-performance brain-computer interfaces (BCIs) \cite {ref4, ref5, ref6}. Central to this approach is the decoding of neural signals, and among these, the LFP offers a uniquely powerful data source. Unlike the sparse firing patterns of individual neurons, the LFP represents the synchronized synaptic activity of thousands of neurons, offering a robust signal for single-trial classification \cite{ref29, ref30}.  

LFP recordings approach overcomes the limitations of non-invasive methods in two critical ways. First, direct LFP recordings offer superior spatial resolution, which eliminates contamination from non-neural sources \cite{ref30, ref31}. Second, these recordings provide a high signal bandwidth that allows the analysis of high-frequency oscillations known to be important for olfactory processing. By capturing this clean and full spectrum of neural activity, LFP recordings offer a powerful and decodable feature source for the specific task of binary odor detection, potentially overcoming the limitations of traditional sensors \cite{ref17}.

Previous studies on the decoding of odor presence from neural signals have generally focused on non-invasive methods, which suffer from a critical loss of high-fidelity spectral information \cite {ref9, ref10, ref11}. Scalp electroencephalography (EEG), for instance, struggles with a low signal-to-noise ratio (SNR) due to the deep cortical source of olfactory signals \cite{ref12, ref13}. Consequently, achieving successful classification with these non-invasive signals requires methods that are impractical for real-world applications. For example, many studies rely on averaging responses in dozens of trials, making real-time detection impossible \cite{ref14, ref15}. The challenge of non-invasive decoding is illustrated in a recent study by Rajabi et al., who targeted the Olfactory Bulb (OB) activity non-invasively with an electrobulbogram (EBG) and a 1D CNN.  Their model achieved an Area Under the Curve (AUC) of 0.58, indicating performance close to the chance level for single-trial classification \cite{ref16}. 

We introduce an ensemble of complementary convolutional neural networks, combining an AttentionCNN and a ResCNN, to determine odor presence from 32-channel extracellular LFP recordings. We propose that by shifting from low-fidelity, non-invasive recordings to high-fidelity, direct LFP signals, we can bridge the performance gap. In this paper, thus,  we propose two central hypotheses. First, we hypothesize that the spectro-temporal features within single-trial LFPs provide sufficient information to robustly and accurately classify odor-presence versus odor-absence conditions. Secondly, we hypothesize that neural activity from the olfactory bulb alone is sufficient for the odor presence detection task, without requiring signals from downstream processing areas like the piriform cortex (PCx). By developing an ensemble of deep convolutional neural networks to decode these LFP signals, we demonstrate a significant leap in performance, thereby establishing the feasibility of using LFP recordings as a powerful foundation for odor sensing.

\section{Material and Methods}
In this section, we start by detailing the pre-processing steps applied to LFP signals. Then, we introduce two core architectures, including AttentionCNN and ResCNN. Finally, we explain the procedures for training and ensembling these models, along with the evaluation metrics.

\begin{figure}[!htbp]
  \centering
  \resizebox{\columnwidth}{!}{%
    \begin{tikzpicture}[
      node distance=0.8cm and 1.2cm,
      >=Stealth,
      every node/.style={font=\scriptsize},
      block/.style={
        rectangle, draw=black,
        fill=blue!10,
        rounded corners=3pt,
        text width=2.2cm,
        align=center,
        minimum height=0.7cm
      },
      arrow/.style={
        ->, thick, shorten >=1pt
      }
    ]
      \node[block] (data)  {Data Source\\\& Experimental Context};
      \node[block,right=of data] (prep) {Preprocessing};
      \node[block,right=of prep] (net)  {Network Architecture};

      \node[block,below=of prep] (ens)   {Ensemble Strategy};
      \node[block,left=of ens] (eval)    {Evaluation Metrics};
      \node[block,right=of ens] (train)  {Training Procedure};

      \draw[arrow] (data) -- (prep);
      \draw[arrow] (prep) -- (net);
      \draw[arrow] (net)  -- (train);
      \draw[arrow] (train) -- (ens);
      \draw[arrow] (ens)  -- (eval);
    \end{tikzpicture}%
  }
  \caption{Block diagram illustrating the methods pipeline.}
  \label{fig:method-pipeline}
\end{figure}
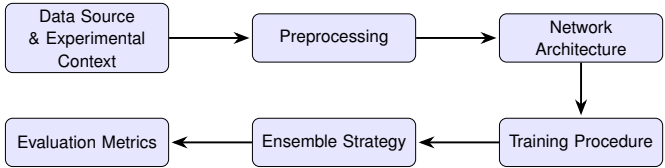

\subsection{Data Source and Experimental Context}
Our analysis uses  \textit{pcx-1} dataset \cite{ref18,ref19} prepared by Bolding and Franks (2018) and is publicly available on CRCNS.org (\href{https://doi.org/10.6080/K00C4SZB}{doi:10.6080/K00C4SZB}). It contains simultaneous extracellular recordings of mice OB and PCx, acquired with 32‐channel NeuroNexus Poly3 silicon probes at a sampling rate of 30\,kHz, together with respiration traces sampled at 2\,kHz. Although both OB and PCx signals are provided, in this study we analyze the OB recordings. The recordings span 2,349 trials from seven awake, head‐fixed mice. Odor stimuli (ethyl butyrate, isoamyl acetate, 2‐hexanone, hexanal, ethyl acetate, and ethyl tiglate) were each delivered at 0.3\,\% v/v; mineral oil served as the baseline.

\subsection{Pre-processing}
A systematic pre-processing pipeline was implemented to prepare the raw LFP signals for analysis. The neural recordings, initially sampled at 30 kHz, were loaded and the dataset was balanced by class using undersampling to mitigate potential biases. Each channel was then subjected to   a fifth-order Butterworth bandpass filter (0.5–100 Hz) for anti-aliasing and noise reduction purposes. This range was chosen to preserve odor-relevant signals while excluding high-frequency noise \cite{ref24,ref25}. Subsequently, the signals were downsampled by a factor of 30 to a new 1 kHz sampling rate, resulting in 2000 samples per trial. This downsampling was performed without introducing aliasing artifacts, as the filter’s 100 Hz cutoff is below the new Nyquist frequency of 500 Hz. Following the pre-processing, spectral features were extracted and normalized. First, power spectral densities (PSDs) were computed for each channel using Welch’s method with a 256-point Hann window and 50.0\% overlap \cite{ref27}.  The resulting spectra were then normalized using a RobustScaler by subtracting the median and dividing it by the inter-quartile range to reduce the impact of  spectral outliers \cite{ref28}.

\subsection{Network Architecture}
\textbf{AttentionCNN.}
AttentionCNN is a well-known architecture for processing complex feature sets. Given the richness and variability of odor-evoked spectral activity, we chose this model to focus on the most discriminative temporal patterns. Our implementation begins with a stand-alone max-pooling layer (stride 2), which is followed by two identical convolutional blocks. Each block consists of the sequence [Conv1D → BatchNorm → ReLU → MaxPool (stride 4)], and these operations collectively expand the output to 128 channels. We then apply three parallel Conv1D branches (kernel sizes 1, 3, 5; 64 filters each), concatenate to form 192 channels, and then recalibrate via a squeeze-and-excitation channel attention module \cite{ref20}, followed by spatial attention. A global average pooling layer reduces the temporal dimension, producing a 192-dimensional vector that passes through dropout (p=0.3), a 256-unit ReLU fully connected layer, dropout (p=0.5), and a final linear classifier for binary odor detection.

\textbf{ResCNN.}
ResCNN builds on residual blocks to enable very deep networks, improving gradient flow and feature reuse. This makes it ideal for capturing hierarchical temporal features in LFP spectra. Our ResCNN adaptation starts with max-pooling (stride 2), a Conv1D layer (kernel 7, stride 2) with BatchNorm and ReLU, and another max-pool (stride 4). Three residual blocks (64 channels each) then refine features, followed by a Conv1D (kernel 3, stride 1 → BatchNorm → ReLU) to expand to 128 channels. After a final max-pool (stride 2) and two more 128-channel residual blocks, global average pooling condenses the temporal dimension. A dropout layer (p=0.4) precedes the final linear classifier, producing robust single-trial odor presence decisions.

\subsection{Training Procedure}
To train and evaluate our models, we employed a five-fold cross-validation scheme on the 2,349 trials. For each fold, the data was partitioned into a training set (80.0\%) and a test set (20.0\%), with 10\% of the training set being used for the validation.  We optimized network parameters using the \textit{AdamW} optimizer (initial learning rate $5\times10^{-4}$, weight decay $1\times10^{-4}$) \cite{ref21}. The learning rate schedule was varied depending on the model being evaluated: for ResCNN and AttentionCNN, the learning rate followed a Cosine Warm Restarts schedule ($T_0 = 10$ epochs, $T_{\mathrm{mult}} = 2$). For the ensemble evaluation, the models within each fold were trained using a One-Cycle policy \cite{ref22}.  Early stopping was applied in each run, with a patience of 15–20 epochs and a minimum required improvement ($\Delta$) of 0.001 in the validation loss. The model checkpoint with the best validation loss was then used for the final testing.

\subsection{Ensemble Strategy}
To combine the complementary outputs of AttentionCNN and ResCNN without additional training, we employ a late fusion of their softmax probabilities. For a given trial \(x\), let
\[
p_{\mathrm{Res}}(x)=\mathrm{softmax}\bigl(m_{\mathrm{Res}}(x)\bigr)\quad  \text{and} \quad  
p_{\mathrm{Att}}(x)=\mathrm{softmax}\bigl(m_{\mathrm{Att}}(x)\bigr).
\]
We compute the ensemble probability as the arithmetic mean,
\[
p_{\mathrm{ens}}(x)=\frac{p_{\mathrm{Res}}(x)+p_{\mathrm{Att}}(x)}{2},
\]
and assign an “odor” label if \(p_{\mathrm{ens}}(x)>0.5\). This simple fusion does not require additional parameters and incurs minimal inference overhead. Under five-fold cross-validation, this probability averaging was performed within each fold: For each of the five splits, an AttentionCNN and a ResCNN were trained on the training portion, and their ensembled predictions were evaluated on the held-out test portion. The final reported metrics are the mean and standard deviation of the performance across these five folds.

\subsection{Evaluation Metrics}
We evaluated classification performance using accuracy, AUC, precision, recall (sensitivity), specificity, and F1 score, and we used confusion matrices to clarify detailed error distributions. All metrics were calculated within each fold of a five‐fold cross‐validation and reported as mean±standard deviation. Furthermore, to verify the reliability of the probability estimates, we compared the model output with actual labels for each test trial to assess the calibration (for example, a predicted odor probability $\sim$ 80.0\% typically indicated actual odor trials $\sim$ 80.0\%).

\section{Results}

\subsection{Performance Metrics Distribution Across Models }
Table~\ref{tab:model_comparison} summarizes five-fold cross-validation results. AttentionCNN reached 79.1\% accuracy, while ResCNN achieved 86.6\%, a +7.5\,pp (percentage point) gain ($\Delta$A). The ensemble matched this peak accuracy (86.6\%, +7.5\,pp vs.\ AttentionCNN). Precision and specificity followed similar patterns, with ResCNN showing higher precision than AttentionCNN, and the ensemble trading a small amount of precision for higher specificity. These results indicate that (i) residual connections yield stronger single-trial detection than attention alone, and (ii) a simple late fusion maintains  accuracy gain while improving robustness.

\begin{table}[!htbp]
\caption{
Performance comparison of AttentionCNN, ResCNN, and Ensemble Models (mean $\pm$ SD over five folds). 
\textbf{Abbreviations:} Acc.\% = Accuracy; F1\% = F1 score; AUC = Area Under the Receiver Operating Characteristic Curve; Sens.\% = Sensitivity; Spec.\% = Specificity.
}

\centering
\resizebox{\columnwidth}{!}{
\scriptsize
\setlength{\tabcolsep}{3pt}
\begin{tabular}{|l|c|c|c|c|c|}
\hline
\textbf{Model} 
  & \textbf{Acc. (\%)} 
  & \textbf{F1 (\%)} 
  & \textbf{AUC} 
  & \textbf{Sens. (\%)} 
  & \textbf{Spec. (\%)} \\ 
\hline
AttentionCNN 
  & 79.1 $\pm$ 1.5
  & 76.3 $\pm$ 2.1
  & 0.8675 $\pm$ 0.0150
  & 76.0 $\pm$ 2.5
  & 92.2 $\pm$ 1.8 \\ 
\hline
ResCNN       
  & 86.6 $\pm$ 1.2
  & 85.4 $\pm$ 1.3
  & 0.9211 $\pm$ 0.0110
  & 87.0 $\pm$ 1.9
  & 86.0 $\pm$ 2.2 \\ 
\hline
Ensemble     
  & 86.6 $\pm$ 1.1
  & 81.0 $\pm$ 1.8
  & 0.9247 $\pm$ 0.0095
  & 84.0 $\pm$ 2.0
  & 90.0 $\pm$ 1.7 \\ 
\hline
\end{tabular}
}
\label{tab:model_comparison}
\end{table}

\subsection{Learned Feature Representation}
\begin{figure}[!htbp]
\centering
\includegraphics[width=3in]{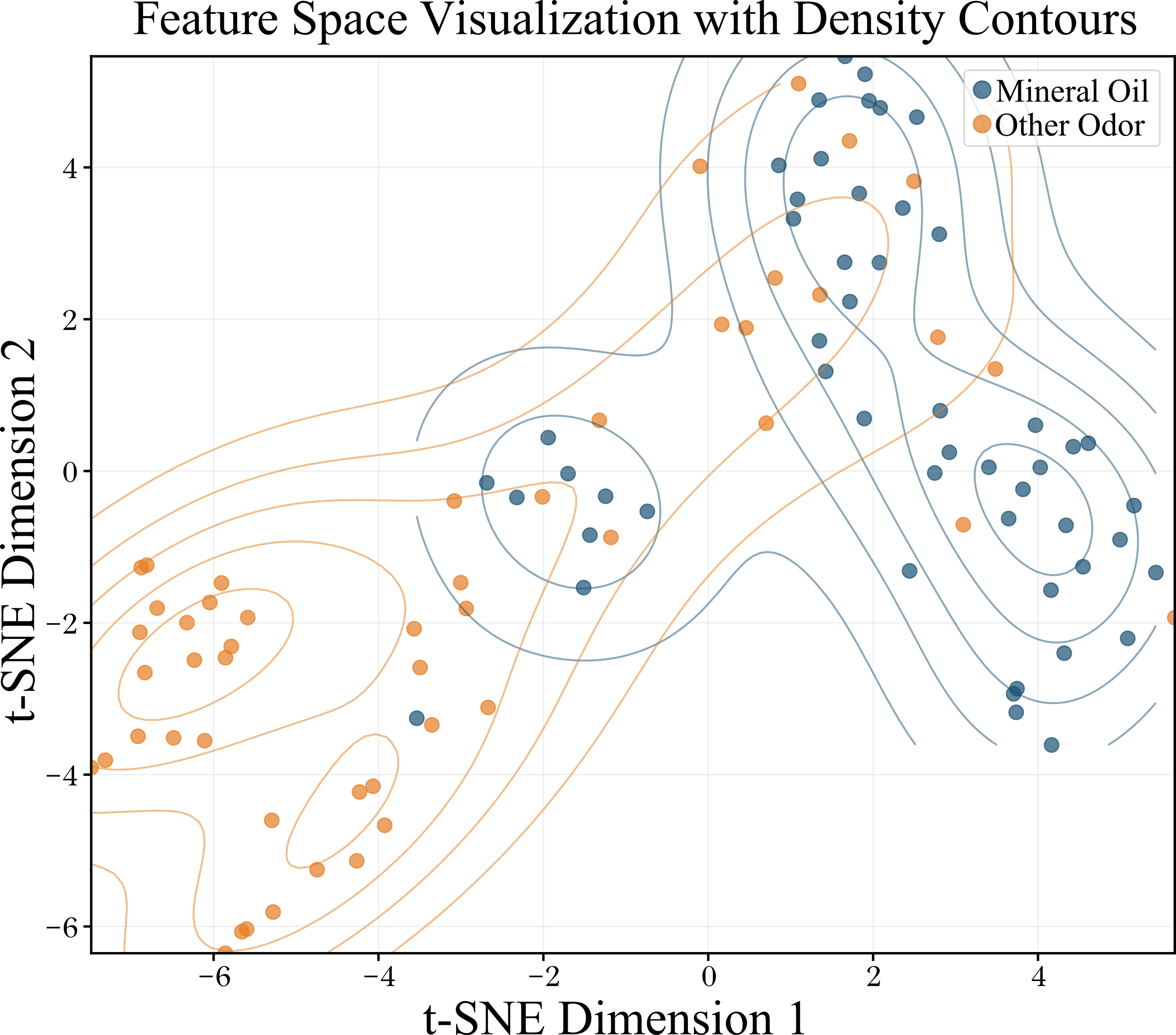}
\caption{t-SNE embedding of the learned features. Distinct clusters for odor (orange) and blank (blue) trials indicate robust separation.}
\label{fig:tsne}
\end{figure}

Beyond quantitative metrics, we validated that our model also learned biologically significant features. Figure~\ref{fig:tsne} visualizes the feature space learned by the network using t-SNE. The clear partitioning between odor (orange) and blank (blue) trials, which form two distinct clusters with minimal overlap, serves as visual confirmation of the model's ability to learn a discriminative feature space. This shows that the spectro-temporal differences between the odor-evoked and baseline neural states are consistent enough to be mapped where they are easily distinguished.

\subsection{Ensemble Prediction Confidence}
\begin{figure}[!htbp]
\centering
\includegraphics[width=3in]{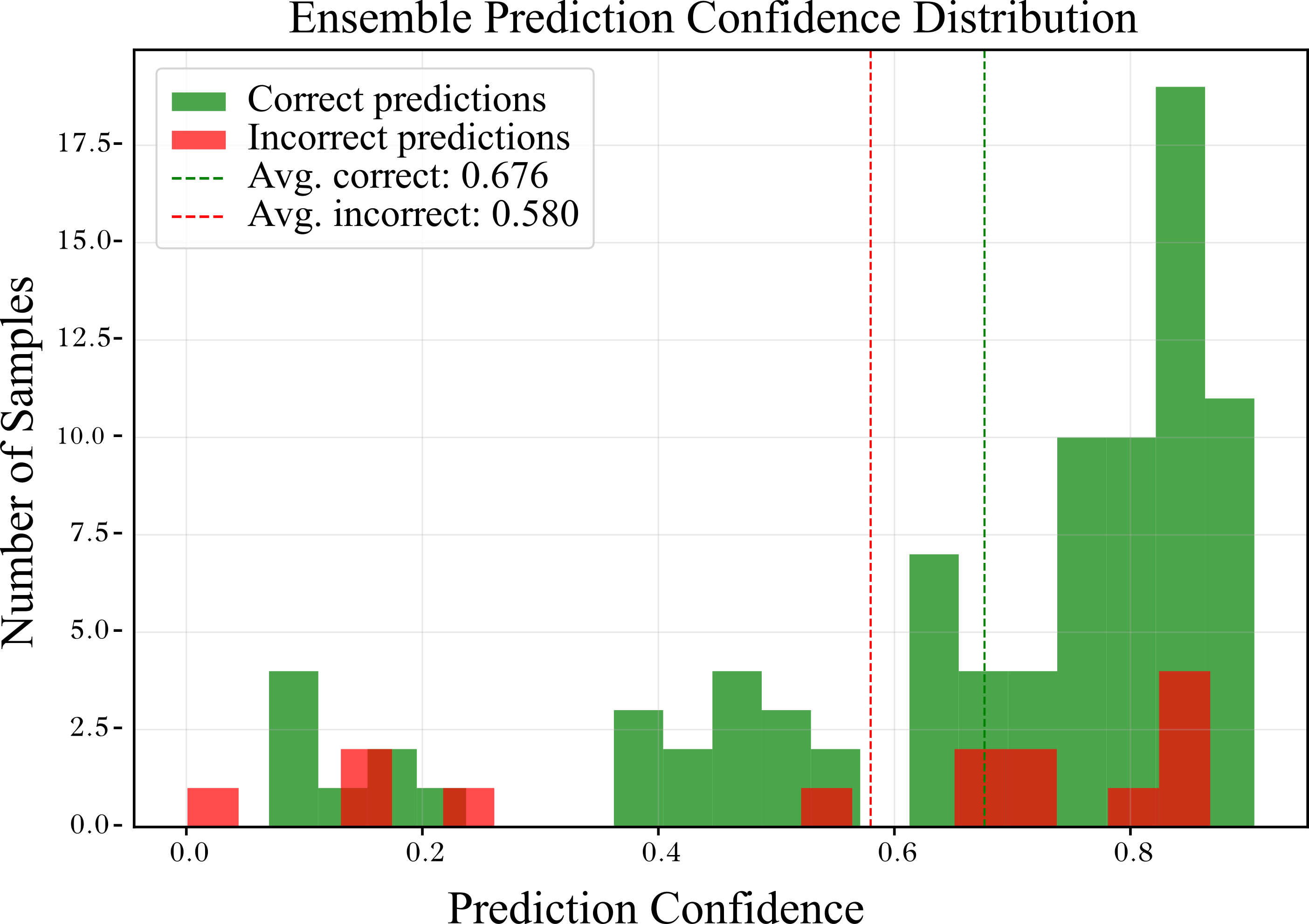}
\caption{Histogram of ensemble prediction confidence for individual test trials, with correct predictions shown in green and incorrect predictions in red. The dashed lines show the average confidence for each group.}
\label{fig:confidence}
\end{figure}

Figure~\ref{fig:confidence} displays the distribution of ensemble prediction confidence for correct and incorrect classifications. The horizontal axis shows the confidence level, and the vertical axis indicates the frequency of trials within each confidence bin. In particular, the green bars correspond to the correct predictions and have higher confidence values (mean confidence $\approx 0.676$), while the red bars (incorrect predictions) are distributed more broadly and have a lower mean confidence ($\approx 0.580$). This suggests that the ensemble classifier is well calibrated as it expresses higher confidence on correct predictions and lower confidence on misclassifications.

\section{Discussion and Conclusion}
\subsection{Overcoming Prior Limitations}
The findings presented in this study offer two primary contributions to the field of olfactory decoding. The first contribution is related to the methodological advance: we establish the feasibility of an accurate, single-trial odor detection model using deep learning on spectral features. This validates our first hypothesis that these signals contain sufficient information for robust classification without averaging. The second contribution is related to the neurological insight: we demonstrate that this performance can be achieved using signals from the olfactory bulb alone. This supports our second hypothesis that the initial stages of olfactory processing are sufficient for the fundamental task of presence detection, without requiring contributions from higher cortical regions.

Prior non-invasive work on human olfactory registration has shown lower performance. Rajabi et al. evaluated logistic regression and an end-to-end 1D ResNet on EEG and EBG signals, finding that their linear baseline remained at AUC $\approx$ 50.0\%, and their ResNet-1D could push AUC into the high-50.0\% range (e.g., 56.6\% for scalp-EBG, 58.0\% for EEG) \cite{ref16}. The difficulty of cross-subject generalization with EEG is further shown by the work of Ezzatdoost et al., who achieved 64.3\% accuracy on a more complex four-odor identification task using handcrafted nonlinear features \cite{ref12}. This demonstrates that the current non-invasive recordings lack sufficient SNR for accurate single-trial odor detection and highlights the advantage of LFP signals.

Table~\ref{tab:gap_analysis} quantifies this advancement, showing that our direct olfactory bulb recordings overcome the fundamental SNR constraints identified by previous researchers. The 30-35\% improvement in AUC validates our central hypothesis that invasive neural access provides the signal fidelity necessary for robust single-trial odor detection.

\begin{table}[!htbp]
\caption{Performance Comparison: Overcoming Prior Limitations}
\centering
\resizebox{\columnwidth}{!}{
\setlength{\tabcolsep}{3pt}
\begin{tabular}{|l|c|c|c|c|}
\hline
\textbf{Study} & \textbf{Signal Type} & \textbf{AUC} & \textbf{Accuracy} & \textbf{Task} \\
\hline
Rajabi et al. & EEG & 0.5800 & 58.0\% & Odor vs. Blank \\
\hline
Rajabi et al. & EBG & 0.5700 & 57.0\% & Odor vs. Blank \\
\hline
Ezzatdoost et al. & EEG & N/A & 64.3\% & 4-Odor Identity (Cross-Subject) \\
\hline
\textbf{Our Work} & \textbf{LFP} & \textbf{0.9247} & \textbf{86.6\%} & \textbf{Odor vs. Blank} \\
\hline
\end{tabular}
}
\label{tab:gap_analysis}
\end{table}

\subsection{Methodological Contributions}
This work establishes a reliable neural-based binary odor detection system, providing both methodological advances and biological insights. Our ensemble approach effectively combines complementary CNN architectures to achieve reliable performance. The clear separation observed in our visualization of t-SNE (Figure~\ref{fig:tsne}) and the high confidence levels for correct predictions (Figure~\ref{fig:confidence}) further validate that our models have learned biologically meaningful features that capture fundamental differences between odor-presence and odor-absence neural states. Compared to prior EEG-based studies, our LFP approach offers higher spatial resolution and direct access to OB circuitry, enabling the extraction of rich spectro-temporal signatures that were previously inaccessible.

\subsection{Limitations and Future Directions}
Our evaluation focuses on monomolecular odorants at a fixed concentration (0.3\% v/v) in head-fixed mice, leaving important questions about generalization to complex mixtures, varying concentrations, and naturalistic behavioral conditions. The invasive nature of LFP recordings also limits immediate translational applications, though our findings provide crucial validation of neural-based odor detection principles. Future work will systematically extend our approach to diverse odor mixtures and concentration ranges, validate its performance in freely moving animals, and investigate non-invasive recording modalities to assess translational feasibility.

\section*{Acknowledgment}
\addcontentsline{toc}{section}{Acknowledgment}
\scriptsize
The research leading to these results has received funding from National Authority TUBITAK with project ID 123E520. We  thank Kevin A. Bolding and Kevin M. Franks for their publicly available dataset.


\normalsize


\begin{thebibliography}{24}

\bibitem{ref1}
T. Sanislav, G. D. Mois, S. Zeadally, S. Folea, T. C. Radoni, and E. A. Al-Suhaimi, "A Comprehensive Review on Sensor-Based Electronic Nose for Food Quality and Safety," \textit{Sensors}, vol. 25, no. 14, p. 4437, Jul. 2025, doi: 10.3390/s25144437.

\bibitem{ref2}
C. Kim \textit{et al.}, "Artificial olfactory sensor technology that mimics the olfactory mechanism: a comprehensive review," \textit{Biomaterials Research}, vol. 26, no. 1, p. 40, Aug. 2022, doi: 10.1186/s40824-022-00287-1.

\bibitem{ref3}
H. Deng, Z. Chen, P. Feng, L. Tian, H. Zong, and T. Nakamoto, "Recent Advances and Applications of Odor Biosensors," \textit{Electronics}, vol. 14, no. 9, p. 1852, May 2025, doi: 10.3390/electronics14091852.

\bibitem{ref4}
E. Shor \textit{et al.}, "Sensitive and robust chemical detection using an olfactory brain-computer interface," \textit{Biosensors and Bioelectronics}, vol. 195, p. 113664, Jan. 2022, doi: 10.1016/j.bios.2021.113664.

\bibitem{ref5}
Q. Lu, M. Yi, and J. Jiang, "Bioelectronic nose for ultratrace odor detection via brain-computer interface with olfactory bulb electrode arrays," \textit{Biosensors and Bioelectronics}, vol. 285, p. 117585, May 2025, doi: 10.1016/j.bios.2025.117585.

\bibitem{ref6}
C. Qin \textit{et al.}, "Artificial Olfactory Biohybrid System: An Evolving Sense of Smell," \textit{Advanced Science}, vol. 10, no. 5, p. 2204726, Jan. 2023, doi: 10.1002/advs.202204726.

\bibitem{ref7}
R. Haddad, T. Weiss, R. Khan, B. Nadler, N. Mandairon, et al.,
"Global features of neural activity in the olfactory system form a parallel code that predicts olfactory behavior and perception," \textit{J. Neurosci.}, vol. 30, no. 27, pp. 9017--9027, 2010.

\bibitem{ref8}
R. W. Friedrich and G. Laurent,
"Dynamic optimization of odor representations by slow temporal patterning of mitral cell activity," \textit{Science}, vol. 291, no. 5505, pp. 889--894, 2001.

\bibitem{ref9}
P. Arpaia \textit{et al.}, "Assessment and Scientific Progresses in the Analysis of Olfactory Evoked Potentials," \textit{Bioengineering}, vol. 9, no. 6, p. 252, Jun. 2022, doi: 10.3390/bioengineering9060252.

\bibitem{ref10}
I. Ninenko, D. F. Kleeva, N. Bukreev, and M. A. Lebedev, "An experimental paradigm for studying EEG correlates of olfactory discrimination," \textit{Frontiers in Human Neuroscience}, vol. 17, p. 1117801, May 2023, doi: 10.3389/fnhum.2023.1117801.

\bibitem{ref11}
B. Iravani \textit{et al.}, "Non-invasive recording from the human olfactory bulb," \textit{Nature Communications}, vol. 11, no. 1, p. 648, Jan. 2020, doi: 10.1038/s41467-020-14520-9.

\bibitem{ref12}
K.~Ezzatdoost, H.~Hojjati, and H.~Aghajan, “Decoding olfactory stimuli in EEG data using nonlinear features: A pilot study,” 
\emph{Journal of Neuroscience Methods}, vol.~341, p.~108780, May 2020, doi:10.1016/j.jneumeth.2020.108780.

\bibitem{ref13}
N. I. Abbasi, R. Bose, A. Bezerianos, N. V. Thakor, and A. Dragomir, 
“EEG-based classification of olfactory response to pleasant stimuli,” 
in Proc. 41st Annu. Int. Conf. IEEE Eng. Med. Biol. Soc. (EMBC), 2019, pp. 5160–5163, doi:10.1109/EMBC.2019.8857673.

\bibitem{ref14}
H.-R. Hou, R.-X. Han, X.-N. Zhang, and Q.-H. Meng, "Pleasantness Recognition Induced by Different Odor Concentrations Using Olfactory Electroencephalogram Signals," \textit{Sensors}, vol. 22, no. 22, p. 8808, Nov. 2022, doi: 10.3390/s22228808.

\bibitem{ref15}
C. Huart, V. Legrain, T. Hummel, P. Rombaux, and A. Mouraux, "Time-Frequency Analysis of Chemosensory Event-Related Potentials to Characterize the Cortical Representation of Odors in Humans," \textit{PLoS ONE}, vol. 7, no. 3, p. e33221, Mar. 2012, doi: 10.1371/journal.pone.0033221.

\bibitem{ref16}
N. Rajabi, I. Zanettin, and A. H. Ribeiro, "Exploring the feasibility of olfactory brain–computer interfaces," \textit{Sci. Rep.}, vol. 15, Art. no. 18404, Jan. 2025, doi: 10.1038/s41598-025-01488-z.

\bibitem{ref17}
Q. Yang \textit{et al.}, "Smell-induced gamma oscillations in human olfactory cortex are required for accurate perception of odor identity," \textit{PLoS Biol.}, vol. 20, no. 1, p. e3001509, Jan. 2022, doi: 10.1371/journal.pbio.3001509.

\bibitem{ref18}
K. A. Bolding and K. M. Franks,
"Recurrent cortical circuits implement concentration-invariant odor coding," \textit{Science}, vol. 361, no. 6407, p. eaat6904, 2018.

\bibitem{ref19}
K. A. Bolding and K. M. Franks,
"Simultaneous extracellular recordings from mouse olfactory bulb (OB) and piriform cortex (PCx) in response to odor stimuli," CRCNS.org, 2018.

\bibitem{ref20}
J. Hu, L. Shen, and G. Sun,
"Squeeze-and-Excitation Networks," in \textit{Proc. IEEE Conf. Comput. Vis. Pattern Recognit.}, 2018, pp. 7132--7141.

\bibitem{ref21}
D. P. Kingma and J. Ba,
"Adam: A Method for Stochastic Optimization," in \textit{Proc. Int. Conf. Learn. Represent.}, 2014.

\bibitem{ref22}
L.~N. Smith, “A disciplined approach to neural network hyper-parameters: Part 1—learning rate, batch size, momentum, and weight decay,” \emph{arXiv preprint arXiv:1803.09820}, 2018.

\bibitem{ref23}
H. Zhang, M. Cisse, Y. N. Dauphin, and D. Lopez-Paz,
"mixup: Beyond Empirical Risk Minimization," in \textit{Proc. Int. Conf. Learn. Represent.}, 2018.

\bibitem{ref24}
L. van der Maaten and G. Hinton,
"Visualizing Data using t-SNE," \textit{J. Mach. Learn. Res.}, vol. 9, pp. 2579--2605, 2008.

\bibitem{ref25}
G. Lepousez and P.-M. Lledo, "Odor Discrimination Requires Proper Olfactory Fast Oscillations in Awake Mice," \textit{Neuron}, vol. 80, pp. 1010--1024, Nov. 20, 2013, doi: 10.1016/j.neuron.2013.07.025.

\bibitem{ref26}
L. M. Kay, "Two Species of Gamma Oscillations in the Olfactory Bulb: Dependence on Behavioral State and Synaptic Interactions," \textit{J. Integr. Neurosci.}, vol. 2, no. 1, pp. 31--44, 2003.

\bibitem{ref27} Welch, P. D. (1967). The use of fast Fourier transform for the estimation of power spectra: A method based on time averaging over short, modified periodograms. \emph{IEEE Trans. Audio Electroacoust.}, AU-15(2), 70–73.

\bibitem{ref28}
P. J. Rousseeuw and C. Croux, "Alternatives to the Median Absolute Deviation," \textit{Journal of the American Statistical Association}, vol. 88, no. 424, pp. 1273–1283, Dec. 1993, doi: 10.2307/2291267.

\bibitem{ref29}
G. T. Einevoll, C. Kayser, N. K. Logothetis, and S. Panzeri, "Modelling and analysis of local field potentials for studying the function of cortical circuits," \textit{Nature Reviews Neuroscience}, vol. 14, no. 11, pp. 770--785, Nov. 2013, doi: 10.1038/nrn3599.

\bibitem{ref30}
G. Buzsáki, C. A. Anastassiou, and C. Koch, "The origin of extracellular fields and currents — EEG, ECoG, LFP and spikes," \textit{Nature Reviews Neuroscience}, vol. 13, no. 6, pp. 407--420, Jun. 2012, doi: 10.1038/nrn3241.

\bibitem{ref31}
S. Katzner, I. Nauhaus, A. Benucci, V. Bonin, D. L. Ringach, and M. Carandini, "Local origin of field potentials in visual cortex," \textit{Neuron}, vol. 61, no. 1, pp. 35--41, Jan. 2009, doi: 10.1016/j.neuron.2008.11.016.


\end{thebibliography}
\end{document}